\documentclass{article}

\usepackage{microtype}
\usepackage{graphicx}
\usepackage{subfigure}
\usepackage{booktabs}
\usepackage{amssymb}
\usepackage{amsmath}
\usepackage{amsthm}
\usepackage{amsfonts}
\usepackage{dsfont}
\usepackage[utf8]{inputenc}
\usepackage[T1]{fontenc}
\usepackage{bbm}
\usepackage{multirow}
\usepackage{tabularx}
\usepackage{arydshln}
\usepackage[hang,flushmargin]{footmisc}
\usepackage{hyperref}
\usepackage{caption}
\usepackage{xurl}

\usepackage[accepted]{icml2021}

\icmltitlerunning{Representation Learning by Ranking across Multiple Tasks}

\begin{document}

\twocolumn[
\icmltitle{Representation Learning by Ranking across Multiple Tasks}

\vspace{0.2cm}

\centerline{\textbf{Lifeng Gu}} 
\centerline{Tianjin University}
\centerline{\texttt{gulifeng666@163.com}}

\vskip 0.3in
]


\begin{abstract}
In recent years, representation learning has become the research focus of the machine learning community. Large-scale neural networks are a crucial step toward achieving general intelligence, with their success largely attributed to their ability to learn abstract representations of data. Several learning fields are actively discussing how to learn representations, yet there is a lack of a unified perspective. We convert the representation learning problem under different tasks into a ranking problem. By adopting the ranking problem as a unified perspective, representation learning tasks can be solved in a unified manner by optimizing the ranking loss. Experiments under various learning tasks, such as classification, retrieval, multi-label learning, and regression, prove the superiority of the representation learning by ranking framework. Furthermore, experiments under self-supervised learning tasks demonstrate the significant advantage of the ranking framework in processing unsupervised training data, with data augmentation techniques further enhancing its performance.
\end{abstract}

\section{Introduction}
Several works model the representation learning problem as a ranking problem~\cite{varamesh2020self,cakir2019deep}. However, these efforts typically focus on developing ranking-based methods to address specific tasks, like classification. We argue that the representation learning by ranking  framework offers a deeper, more intrinsic connection to representation learning. \par

Representation learning can naturally be framed as a ranking problem. Typically, a neural network first maps inputs into a feature space and then generates prediction labels. The ranking  framework can guide the learning process within this feature space based on the similarity between labels. \par

Consider a model \( f: \mathbb{R}^n \to \mathbb{R}^m \) that maps input samples into an \( m \)-dimensional feature space, with a sample set \( \{x_i \mid i = 1, 2, 3, \ldots, n\} \) and a label set \( \{y_i \mid i = 1, 2, 3, \ldots, n\} \). For any sample \( x_i \), we treat it as a query, with all other samples \( \{x_j \mid j = 1, 2, 3, \ldots, n, j \neq i\} \) as candidates. The model transforms \( x_i \) into the feature \( f(x_i) \) and similarly maps the candidate samples into \( \{f(x_j) \mid j \neq i\} \). Using a predefined similarity function, we compute the similarity set \( \{\text{sim}(f(x_i), f(x_j)) \mid j \neq i\} \). Our objective is to rank this similarity set according to a true order, thereby guiding the model’s learning.

A straightforward solution is to use a ranking loss to optimize the order within the similarity set, guided by the true order computed from the samples' labels.

We utilize an approximate NDCG loss to optimize this order. Experiments across various learning tasks, such as classification, retrieval, multi-label learning, regression, and self-supervised learning, demonstrate the superiority of the approximate NDCG loss. This not only highlights the superiority of the approximate NDCG loss, but also verifies the effectiveness and broad applicability of the representation learning by ranking framework.

\section{Related Works}

\subsection{representation learning}
According to~\cite{bengio2013representation}, representation learning aims to derive data representations that facilitate the extraction of useful information for building classifiers or other predictors. A good representation should also serve as an effective input for supervised predictors. Although numerous fields investigate this problem, a unified perspective remains elusive. We propose categorizing representation learning into supervised, self-supervised, and unsupervised approaches. For instance, supervised ImageNet pre-training models can reduce the complexity of downstream tasks by leveraging representations learned from ImageNet as input. Self-supervised visual representation learning methods, such as SimCLR~\cite{chen2020simple} and BYOL~\cite{grill2020bootstrap}, mitigate the complexity of visual tasks. Unsupervised generative models, including VAEs~\cite{kingma2013auto,oord2017neural} and BiGANs~\cite{donahue2016adversarial,donahue2019large}, enable data generation or disentangle variation factors~\cite{chen2016infogan,dupont2018learning,kim2018disentangling}. Similarly, unsupervised language models like BERT~\cite{devlin2018bert} learn contextual language representations, simplifying downstream language tasks.

\subsection{Deep Metric Learning}
Deep metric learning seeks to develop effective metrics for measuring sample similarity, typically by computing distances between representations. In tasks like retrieval, a good representation inherently implies a robust metric, and vice versa. Popular deep metric learning methods include Contrastive Loss~\cite{hadsell2006dimensionality}, Center Loss~\cite{wen2016discriminative}, N-Pair Loss~\cite{sohn2016improved}, and Circle Loss~\cite{sun2020circle}.\cite{zhu2018} proposed the concept of relation alignment, extending metric learning to multiple tasks, which motivated us to generalize it to broader areas.
\subsection{Self-Supervised Representation Learning}
In recent years, self-supervised learning has gained significant attention for representation learning. Early work, such as Deep InfoMax (DIM)~\cite{hjelm2018learning}, simultaneously estimates and maximizes mutual information between input data and high-level representations, employing adversarial learning to align representations with prior constraints. Contrastive Predictive Coding (CPC)~\cite{oord2018representation} uses a powerful autoregressive model to predict future representations in latent space, introducing the InfoNCE objective, which has since become widely adopted. Contrastive Multiview Coding (CMC)~\cite{tian2020contrastive} builds on the assumption that good representations maintain consistency across perspectives, achieving this by maximizing mutual information between different views of the same sample; more views typically yield better results. 

Tschannen et al.~\cite{tschannen2019mutual} synthesized prior work to analyze the principles of mutual information maximization. They argue that while this approach often enhances downstream task performance, it can occasionally degrade it. They suggest that its success may be attributed to its alignment with metric learning, a proven paradigm for effective representation learning. DeepCluster~\cite{caron2018deep} integrates clustering with representation learning, iteratively assigning cluster labels and using them as pseudo-labels to refine representations. SwAV~\cite{caron2020unsupervised} employs multiple prototypes for clustering, ensuring consistency across data views.

Recent developments have further advanced the field. DINO~\cite{caron2021emerging} introduces self-distillation with information noise, eliminating the need for negative samples while achieving state-of-the-art results. SimSiam~\cite{chen2021exploring} explores a simple siamese architecture, leveraging positive samples alone to learn robust representations. Shen et al.~\cite{shen2020rethinking} investigate the impact of hybrid data augmentation strategies on representation quality, while Tian et al.~\cite{tian2020makes} provide theoretical insights into optimal view selection for maximizing representation quality.
\section{Background}

In this section, we discuss the ranking problem and learning to rank.

\subsection{Ranking}
Ranking and learning to rank are classic problems with extensive research~\cite{xu2008directly,xia2008listwise}. Given an input query, a retrieval system aims to sort and return stored content based on its relevance to the query. The goal of learning to rank is to improve the accuracy of the returned results. One solution is to optimize evaluation metrics. Common evaluation metrics for ranking problems include Precision, Average Precision (AP)~\cite{baeza1999modern}, and Normalized Discounted Cumulative Gain (NDCG)~\cite{jarvelin2002cumulated}; for further details, see~\cite{qin2010general}.

Given a query sample $q$, and a returned sorted sample set $S$, the precision at $k$ is defined as:
\begin{equation}
    \text{Prec}@k = \frac{1}{k}\sum_{j=1}^{k}r_{j}
\end{equation}
where $r_{j}\in\{0,1\}$ indicates the relevance of the $j$-th returned sample to the query sample. If $S_{j}$ is relevant to the query, $r_{j}=1$; otherwise, $r_{j}=0$.

Average Precision (AP) is defined based on precision at $k$ as:
\begin{equation}
    \text{AP} = \frac{1}{N}\sum_{j}{r_{j} \times \text{Prec}@_{j}}
\end{equation}
where $N$ represents the number of samples related to the query sample in the returned sample set $S$.

Recent works~\cite{cakir2019deep,brown2020smooth} have used AP as the optimization target. However, AP is only applicable when the returned samples and the query sample are either correlated or uncorrelated. For multi-label learning tasks, AP is not suitable.

NDCG is an extension of AP that can handle multi-level correlations between returned samples and query samples:
\begin{equation}
    \text{NDCG} = N_{n}^{-1}\sum_{j=1}^{n}g(r_{j})d(j)
\label{eq:nddcg}
\end{equation}
where $n$ represents the size of the sample set $S$, $r_{j}\ge 0$ represents the relevance score between the $j$-th returned sample and the query sample, and $N_{n}$ is a normalization factor. It represents the value of $\sum_{j=1}^{n}g(r_{j})d(j)$ when the returned samples are sorted according to their true relevance to the query sample in descending order. This normalization ensures that the value of NDCG is not greater than 1. $g(r_{j})$ represents the gain function, and $d(j)$ represents the discount function. Following~\cite{qin2010general}, the default settings are: $g(r_{j})=2^{r_{j}}-1$ and $d(j) = 1/\log_{2}(1+j)$. Substituting these values into Equation~\eqref{eq:nddcg}, we obtain:
\begin{equation}
    \text{NDCG} = N_{n}^{-1}\sum_{j=1}^{n}(2^{r_{j}}-1)/\log_{2}(1+j)
\label{eq:new_nddcg}
\end{equation}
\section{A-NDCG}
What defines a model with strong representation capabilities? We hypothesize that for any input sample, the similarity set generated by the model should exhibit a correct order: samples similar to the input should rank higher, while dissimilar ones should rank lower. This assumption has been widely validated in classification tasks, and we seek to extend this reasonable hypothesis to a variety of learning tasks. Leveraging the principles of learning-by-ranking, we optimize a unified ranking objective:
\begin{equation}
\text{Rank}(\text{sim}(f(x_i), f(x_j))) = \text{Rank}(\text{sim}(y_i, y_j))
\end{equation}
where \( j = 1, 2, \ldots, n \) and \( j \neq i \).

Here, \( x_i \) represents any training sample, and \( f(x_i) \) is an \( m \)-dimensional vector representing the feature embedding of \( x_i \). The function \(\text{sim}(f(x_i), f(x_j))\) denotes the similarity between the feature embeddings of samples \( x_i \) and \( x_j \), while \(\text{sim}(y_i, y_j)\) denotes the similarity between their corresponding labels. The objective is to ensure that the ranking of similarities in the feature space matches the ranking of similarities in the label space.

\begin{figure*}[htbp]
\centering
\includegraphics[width=0.95\textwidth]{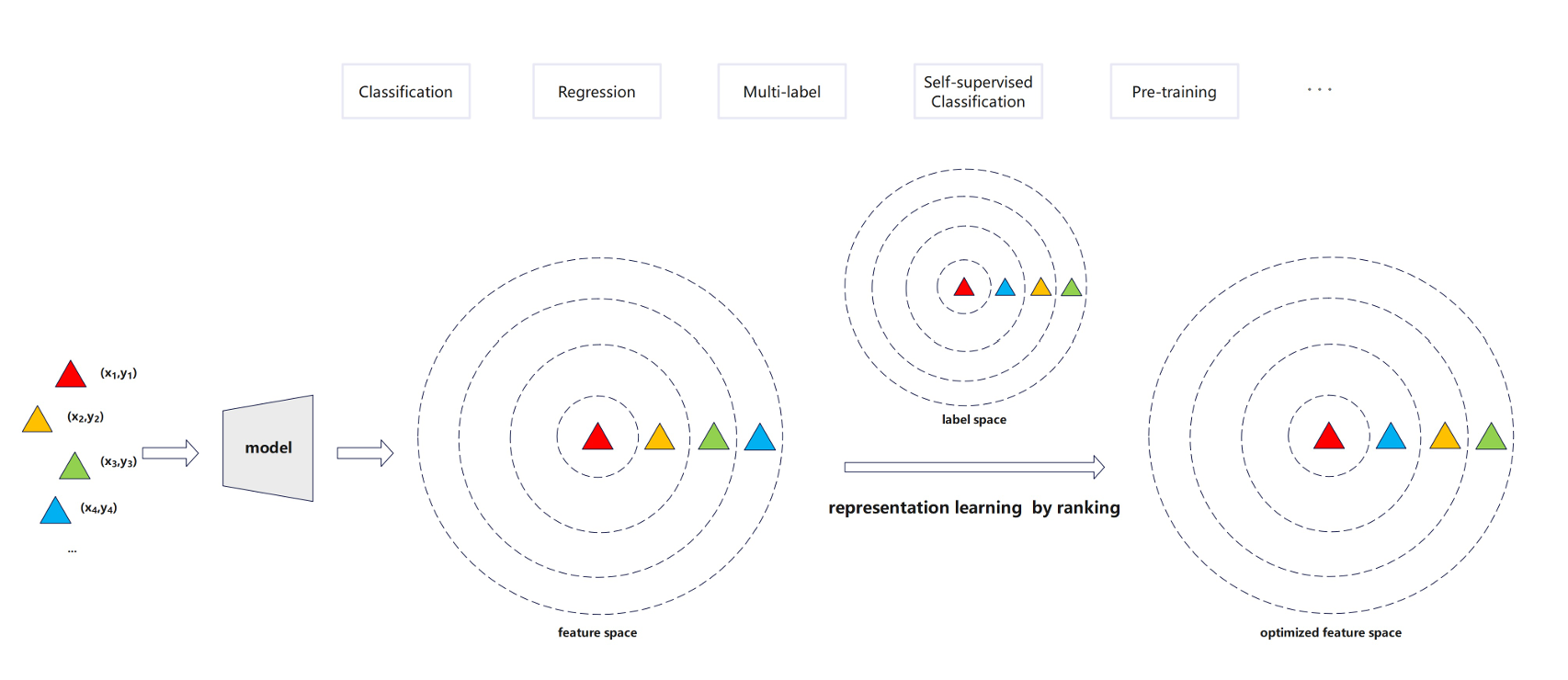}
\caption {Suppose there are four samples, \(x_1, x_2, x_3, x_4\), and their corresponding labels \(y_1, y_2, y_3, y_4\). For the query sample \(x_1\), if in the label space it satisfies \( \text{sim}(y_1, y_4) > \text{sim}(y_1, y_2) > \text{sim}(y_1, y_3) \), then in the feature space, we also hope to satisfy \( \text{sim}(f(x_1), f(x_4)) > \text{sim}(f(x_1), f(x_2)) > \text{sim}(f(x_1), f(x_3)) \), i.e., the order of similarity is preserved. }
\label{fig:rank}
\end {figure*}

The similarity function \(\text{sim}\) varies depending on the learning task. In general, we define the similarity in the feature space as \(\text{sim}(f(x_i), f(x_j)) = f(x_i) \cdot f(x_j)^T\). The similarity in the label space is defined as follows for different tasks:

\begin{itemize}
    \item \textbf{Classification:} \(\text{sim}(y_i, y_j) = \mathbb{I}[y_i = y_j]\), where \(\mathbb{I}\) is the indicator function, which equals 1 if \(y_i = y_j\) and 0 otherwise.
    \item \textbf{Regression:} \(\text{sim}(y_i, y_j) = 1 - \frac{|y_i - y_j|}{\max(|y_i - y_j|)}
\) , representing the normalized absolute difference between the target values.
    \item \textbf{Multi-label Classification:} \(\text{sim}(y_i, y_j) = \text{cosine}(y_i, y_j) = \frac{y_i \cdot y_j^{T}}{\|y_i\| \|y_j\|}\), measuring the cosine similarity between the label vectors.
     \item \textbf{Self-supervised Classification:}
    \(\text{sim}(y_i, y_j)\) represents the similarity of pseudo-labels.
    \item \textbf{Pre-training:} \(\text{sim}(y_i, y_j)\) can represent the word distance within a sentence, or other relevant relationships depending on the pre-training task.
\end{itemize}
Figure\ref {fig:rank} provides an example to illustrate the representation learning by ranking framework.
 \par
 We use the approximate NDCG indicator to optimise the objective . For any sample $x_{i}$ from the sample set as the query sample, approximate NDCG indicator or A-NDCG loss can be formulated: 
\begin{equation}
     L(x)=\sum_{i} N_{i}^{-1}\sum_{j,j\neq i}^{n}(sim(y_{i},y_{j})/log_{2}(1+\pi(x_{i},x_{j}))
\end{equation}
 \begin{eqnarray}  
    \pi(x_{i},x_{j})&=& 1+\sum_{k,k\neq j} {\frac {exp(-\alpha sim_{ijk}
    ))}{1+exp(-\alpha sim_{ijk} )}}\\
    \nonumber
    sim_{ijk} &=& (sim(f(x_{i}),f(x_{j}))-sim(f(x_{i}),f(x_{k})))\\
    \nonumber
\end{eqnarray}
Here, \(\alpha\) is a hyperparameter, and \(N_i^{-1}\) is the normalization term, representing the maximum value of \(\sum_{j=1}^n \operatorname{sim}(y_i, y_j) / \log_2(1 + \pi(x_i, x_j))\) when the order of \(\{\operatorname{sim}(f(x_i), f(x_j)) \mid j \neq i\}\) matches that of \(\{\operatorname{sim}(y_i, y_j) \mid j \neq i\}\).
\par
NDCG is non-differentiable due to the position index \(j\). In A-NDCG, \(\pi(x_i, x_j)\) serves as an approximation of the position index \(j\) used in the NDCG calculation.

The advantages of the approximate NDCG loss are as follows:
\begin{enumerate}
    \item It can naturally incorporate an arbitrary number of perspectives of the training data, thereby relaxing the constraint of using only two perspectives as required by popular contrastive learning algorithms~\cite{chen2020simple,grill2020bootstrap}.
    \item It imposes minimal constraints, focusing solely on the ordering of samples without requiring additional conditions, which facilitates the learning of robust representations.
    \item Compared to contrastive learning methods~\cite{chen2020simple,grill2020bootstrap} and learning-to-rank methods based on optimized average precision~\cite{varamesh2020self}, the approximate NDCG loss is applicable in any scenario where a label similarity set can be obtained. This versatility makes it suitable for a wide range of learning tasks.
    \item Due to its capability to handle diverse label types, label-level data augmentation methods can be applied to the training data to continuously enhance the performance of the approximate NDCG loss, thereby fully leveraging the available training data.
\end{enumerate}
\section{A Unified Perspective on Understanding Modern  Language Models and Classical Methods}
\begin{figure*}[htbp]
\centering
\includegraphics[width=0.75\textwidth]{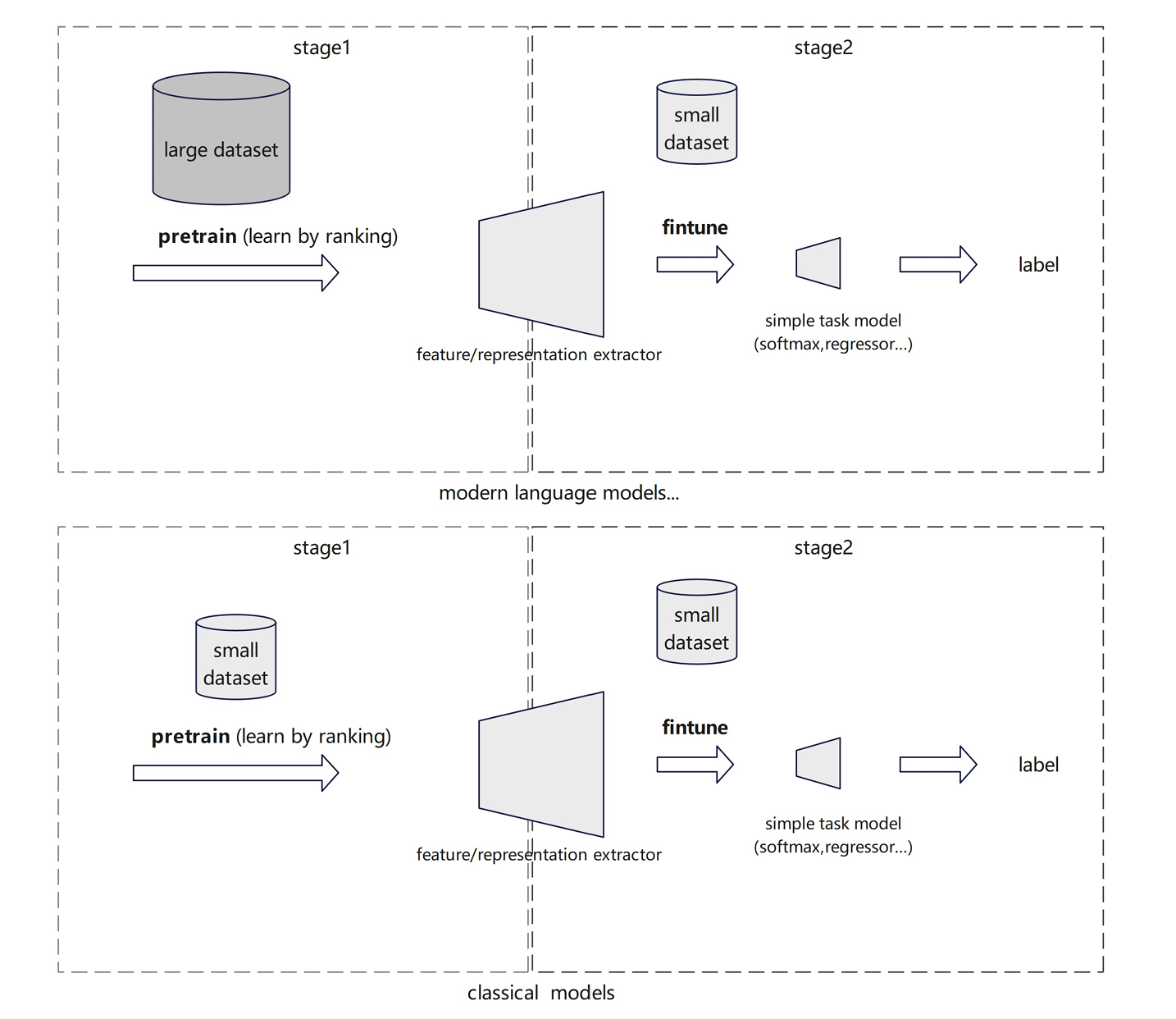}
\caption {a unified perspective on understanding modern  language models and classical methods }
\label{fig:unified}
\end {figure*}

Thanks to the broad applicability of ranking-based representation learning, we can understand modern and classical methods from a unified perspective. Modern  language models typically consist of two stages: pre-training and fine-tuning. Classical methods (such as traditional image classification models) usually do not have a pre-training stage. Through the representation learning by ranking  framework, the learning process of classical methods can also be divided into two stages: the feature learning stage using ranking loss like A-NDCG and the label learning stage using a simple classifier or regressor. This naturally corresponds to the pre-training and fine-tuning stages. The only difference is that the pre-training stage of modern  language models uses a large dataset different from the downstream task, while classical methods use a small dataset that is the same as the downstream task. This unification provides deeper insights into understanding representation learning and offers a broader perspective for designing specific pre-training objectives. By simply identifying ranking relationships, new pre-training objectives can be designed. Figure \ref {fig:unified} illustrates this unification.

\section{Experiment}
In order to evaluate A-NDCG and verify the advantages of learning through a ranking framework, we conducted experiments under a variety of learning tasks. The experiments in this paper include learning representations for various tasks: classification tasks, retrieval tasks, self-supervised tasks, multi-label classification, and regression tasks.

\subsection{Classification Task}
In the classification task, the learned representation should be able to effectively utilize linear classifiers such as softmax to solve classification problems. This paper uses Cross-Entropy loss and its variant~\cite{liu2016large}, as well as the supervised contrastive learning algorithm SupCon~\cite{caron2020unsupervised}, as comparison algorithms. Then, the classification accuracy of a linear softmax classifier on the popular CIFAR-10 and CIFAR-100 datasets is used to evaluate the approximate NDCG loss.

\subsubsection{Implementation Details}
For the approximate NDCG loss, this paper uses the standard residual network, ResNet-50, as the encoder. Following the standard practice~\cite{chen2020simple}, a small projection network composed of a two-layer MLP and ReLU activation function is added after the residual network. We use the standard Adam optimizer~\cite{kingma2014adam}.

\subsubsection{Experimental Results Analysis}
Table \ref{table:classify} shows that the performance of A-NDCG on the CIFAR-10 and CIFAR-100 datasets exceeds that of Cross-Entropy loss and some of its variants, such as Max-Margin~\cite{liu2016large}. It is equivalent to the performance of SupCon~\cite{khosla2020supervised}. Compared to Cross-Entropy loss and Max-Margin~\cite{liu2016large}, A-NDCG can utilize the relationship between sample pairs instead of just the relationship between a single sample and its label. It also relaxes the limitations of SimCLR~\cite{chen2020simple} regarding the number of perspectives and allows for the use of more comparative information between samples.

\subsection{Retrieval Task}
The image retrieval task is a standard evaluation task in the field of deep metric learning. The representation learned for the retrieval task should be able to use linear learners such as k-NN to retrieve samples.

We compare various deep metric learning algorithms~\cite{wang2019multi}. We use the standard dataset CUB-200-2011~\cite{WahCUB_200_2011} to evaluate the retrieval performance of the approximate NDCG loss.

The implementation details are consistent with the previous section.

\subsubsection{Experimental Results Analysis}
Table \ref{table:classify_cub} shows that on the CUB-200-2011 dataset, the performance of A-NDCG exceeds that of many popular deep metric learning methods. In the specific implementation of this paper, the number of positive and negative samples involved in the calculation is the same as in other methods. Compared to other methods, A-NDCG only needs to constrain the order relationship without specifying a specific sample interval, resulting in more robust performance on the test data.
\begin{table*}[htbp]
\caption{Classification accuracy on CIFAR-10 and CIFAR-100 datasets}\label{table:classify}
\vspace{0.5em}\centering
\begin{tabular}{ccccccc}
\toprule[1.5pt]
dataset & SimCLR &  Cross-Entropy & Max-Margin\cite{liu2016large} &SupCon\cite{caron2020unsupervised} & A-NDCG\\
\midrule[1pt]
CIFAR10 &93.6& 95.0& 92.4 &96.0&95.3\\
CIFAR100 &70.7 &75.3 &70.5& 76.5&76.7\\
\bottomrule[1.5pt]
\end{tabular}

\label{table:classfiy_cifar}
\end{table*}
\begin{table}[htbp]
\caption{Retrieval performance on CUB-200-2011 dataset, we use recall to estimate}
\vspace{0.5em}\centering
\begin{tabular}{ccccccc}
\toprule[1.5pt]
Rank@K & 1& 2& 4 &8 & 16&32\\
\midrule[1pt]
Clustering$^{64}$ &48.2&	61.4&	71.8&	81.9&   -&	-\\
ProxyNCA$^{64}$ &49.2	&61.9&	67.9&	72.4&	-&	-\\
Smart Mining$^{64}$&49.8&	62.3&	74.1&	83.3&	-& -	\\
HTL$^{512}$&	57.1&	68.8&	78.7&	86.5&	92.5&	95.5\\
ABIER$^{512}$	&57.5&	68.7&	78.3&	86.2&	91.9&	95.5\\
MS-Loss$^{512}$&57.5&70.3& 80.0& 88.0& 93.2& 96.2\\
A-NDCG$^{512}$&58.3& 70.7& 80.5&88.5&93.8&96.9\\
\bottomrule[1.5pt]
\end{tabular}
\vspace{\baselineskip}
\label{table:classify_cub}
\end{table}
\subsection{Multi-label Learning}
Multi-label learning is a traditional research direction, and there have been quite a lot of research results~\cite{zhang2007ml}, where the representations obtained should be able to reduce the learning difficulty of other linear multi-label learning algorithms. \par
Following \cite{zhu2018}, We use Hamming loss and Jaccard score to evaluate the performance. Hamming loss measures the difference between predicted and true multi-label labels using Hamming distance, where lower values are better. The Jaccard score measures the ratio of the intersection to the union of the predicted and true labels, with higher values indicating better performance.
\subsubsection{Dataset}
We use a variety of popular multi-label datasets \footnote{http://mulan.sourceforge.net/datasets-mlc.html}. Core5k is an image dataset with 5000 images, Scene has more than 2000 images, Medical and Enron are two text data, Medical has 978 samples, and Enron has 1702 samples.
\subsubsection{Evaluation Algorithm}
MLKNN and BRKNN are two distance-based multi-label learning algorithms, we use them to evaluate A-NDCG.
\subsubsection{Experimental Results Analysis}
From the table \ref{table:muti_brknn}, table \ref{table:muti_mlknn}, it can be clearly seen that as the number of iterations increases, the Hamming loss continues to decrease and the Jaccard score increase continually , A-NDCG loss is very effective in learning good representations on multi-label data. A-NDCG loss makes full use of the label information of the multi-label dataset, even for samples with very close multi-label labels. It can give specific optimization goals on how to distinguish them, so that the samples can distinguish samples with similar labels in the feature space, and ultimately reduce the learning difficulty of the linear multi-label learner.
\begin{table*}[htbp]
\caption{classification accuracy on multi-label datasets, we use MLKNN\cite{zhang2007ml} to evaluate.}
\vspace{0.5em}\centering
\begin{tabular}{c|c|ccccccccccc}
\toprule[1.5pt]

dataset&indicator & MLKNN& A-NDCG, epoch:10&20&30&40&50\\
\midrule[1pt]
\multirow{2}{*}{Scene} &{Hamming Loss}& 0.102 &0.095 &  0.088&0.090&0.088&0.093 \\
                        &{Jaccard Score}& 0.610  & 0.698  &0.707  &0.710&0.722&0.715 \\
\multirow{2}{*}{Corel5k} &{Hamming Loss}& 0.012  & 0.011&0.011    & 0.011&0.012&0.012  \\
                        &{Jaccard Score}& 0.094  &0.1116   &0.134 &0.133&0.133&0.125   \\
\multirow{2}{*}{Medical} &{Hamming Loss}& 0.020  & 0.013    &0.013& 0.013 &0.013&0.011 \\
                        &{Jaccard Score}& 0.512  &0.726    & 0.74& 0.74 &0.74&0.739 \\
\multirow{2}{*}{Enron} &{Hamming Loss}& 0.062  &0.05   & 0.05 &0.05&0.05&0.056 \\
                        &{Jaccard Score}& 0.329  &  0.441  &0.449& 0.451 &0.455&0.456 \\
\bottomrule[1.5pt]
\end{tabular}

\label{table:muti_mlknn}
\end{table*}
\begin{table*}[htbp]
\caption{classification accuracy, we use BRKNN~\cite{EleftheriosSpyromitros2008} to evaluate.}
\vspace{0.5em}\centering
\begin{tabular}{c|c|ccccccccccc}
\toprule[1.5pt]
dataset&indicator & BRKNN& A-NDCG, epoch:10&20&30&40&50\\
\midrule[1pt]
\multirow{2}{*}{Scene} &{Hamming Loss}& 0.109 &0.095 &  0.095&0.090&0.089&0.091   \\
                        &{Jaccard Score}& 0.640  & 0.698  &0.720  &0.725&0.726&0.726   \\
\multirow{2}{*}{Corel5k} &{Hamming Loss}& 0.011 & 0.011    &0.011 &0.011&0.011& 0.012  \\
                        &{Jaccard Score}& 0.069 &0.123    &0.140 &0.145&0.149&0.142 \\
\multirow{2}{*}{Medical} &{Hamming Loss}& 0.020  &0.014     &  0.013&0.013 &0.014&0.013 \\
                        &{Jaccard Score}& 0.472  & 0.696  &0.703& 0.709  &0.72&0.723 \\
\multirow{2}{*}{Enron} &{Hamming Loss}& 0.059 &   0.05 & 0.05 & 0.05&0.05&0.052\\
                        &{Jaccard Score}& 0.324  & 0.44  &0.46 &0.46 &0.46&0.472   \\
\bottomrule[1.5pt]
\end{tabular}
\label{table:muti_brknn}
\end{table*}
\subsection{Regression Task}
The regression task is widely used in various fields such as time series forecasting, energy forecasting, and financial market forecasting. The representation learned under the regression task should be able to reduce the learning difficulty of the linear regressor.

We use ridge regression and linear regression methods as evaluation algorithms. We use absolute loss (MAE) and mean square loss (MSE) as evaluation metrics to evaluate A-NDCG. Both metrics measure the error of the regression results, with lower values indicating better performance.

We selected several regression datasets from the UCI dataset repository, including housing price data, wine data, and Parkinson's disease data.

\subsubsection{Experimental Analysis}
The regression task is indeed challenging. Unlike discriminative tasks, it involves continuous labels. However, A-NDCG can still learn good representations in regression datasets, thereby reducing the learning difficulty of regression methods. Some works~\cite{hooshmand2019energy,ye2018novel} have combined pre-training models with prediction tasks. We believe that A-NDCG can also be naturally applied to such tasks.
\begin{table}[htbp]
\caption{result on regression datasets, we use ridge regression method to evaluate.}
\vspace{0.5em}\centering
\begin{tabular}{c|c|ccccccccccc}
\toprule[1.5pt]
dataset&indicator& ridge& A-NDCG\\
\midrule[1pt]
\multirow{2}{*}{parkinsons} &{MSE}& 91.4 & 77.61\\
                        &{MAE}& 7.90 & 7.40 \\

\multirow{2}{*}{housing} &{MSE}& 18.61  &-  \\
                        &{MAE}& 3.40  &-\\
                        \multirow{2}{*}{wine} &{MSE}& 0.62  &     0.59  \\
                        &{MAE}& 0.60  &  0.59 \\

\bottomrule[1.5pt]
\end{tabular}
\vspace{\baselineskip}
\label{table:self_mix_knn}
\end{table}
\begin{table}[htbp]
\caption{result on regression datasets, we use linear regression methods to evaluate.}
\vspace{0.5em}\centering
\begin{tabular}{c|c|ccccccccccc}
\toprule[1.5pt]
dataset&indicator & LR& A-NDCG\\
\midrule[1pt]
\multirow{2}{*}{parkinsons} &{MSE}& 91.42 & 72.72 \\
                        &{MAE }& 7.433 & 7.044 \\
\multirow{2}{*}{housing} &{MSE}& 18.64 &     13.77  \\
                        &{MAE}& 3.398  &  2.95 \\
\multirow{2}{*}{wine} &{MSE}& 0.62 &  0.55  \\
                        &{MAE}& 0.60  & 0.58\\

\bottomrule[1.5pt]
\end{tabular}
\vspace{\baselineskip}
\end{table}
\subsection{Self-supervised Learning Task}
Although there have been works that introduce the idea of learning to rank into the field of self-supervised learning~\cite{varamesh2020self}, the optimization goal of this paper is different. \cite{varamesh2020self} uses average recall as the optimization goal. In self-supervised tasks, the representations learned using unsupervised data should be able to reduce the difficulty of supervised learning, such as improving the classification performance of linear learners.

We conduct experiments on the popular STL-10 dataset~\cite{coates2011analysis}. In this paper, logistic regression and k-nearest neighbor classifiers are used as methods for evaluating representations.

We also use the data augmentation methods mixup~\cite{zhang2017mixup} and cutmix~\cite{yun2019cutmix} to augment the training data at the label level to verify whether A-NDCG can fully utilize the information in the unsupervised data.
\begin{table*}[htbp]
\caption{Classification accuracy on STL-10 dataset, we use logistic regression  to evaluate.}\label{tabl:self}
\vspace{0.5em}\centering
\begin{tabular}{ccccccccccccc}
\toprule[1.5pt]
method & epoch:0 &4&8&12&16&20&24&28&32\\
\midrule[1pt]
SimCLR(train)&70.4&78.2&80.7&82.0&83.5&84.2&82.2&82.9&85.8\\
A-NDCG(train)&77.8&85.0&86.1&86.7&86.6&86.6&87.4&87.0&86.3\\
SimCLR(test)&44.0&49.8&51.8&52.3&51.6&52.7&52.7&52.6&51.7\\
A-NDCG(test)&44.0&49.5&51.9&52.3&51,8&52.1&52.8&50.9&52.4\\
\bottomrule[1.5pt]
\end{tabular}
\vspace{\baselineskip}
\label{table:self-acc}
\end{table*}
\begin{table*}[htbp]
\caption{Classification accuracy on STL-10 dataset, we use knn to evaluate.}
\vspace{0.5em}\centering
\begin{tabular}{ccccccccccccc}
\toprule[1.5pt]
method & epoch:0 &4&8&12&16&20&24&28&32\\
\midrule[1pt]
SimCLR(train)&59.2&62.4&62.9&64.5&64.5&65.0&65.0&65.9&66.1\\
A-NDCG(train)&62.2&66.1&66.6&67.5&68.3&67.4&67.8&67.9&68.9\\
SimCLR(test)&31.5&37.0&39.1&40.0&39.6&39.6&39.6&41.1&41.0\\
A-NDCG(test)&36.3&40.5&42.4&43.0&43.0&43.2&44.0&44.8&44.7\\
\bottomrule[1.5pt]
\end{tabular}
\vspace{\baselineskip}
\label{table:self-knn}
\end{table*}
\begin{table*}[htbp]
\caption{Classification accuracy on mixed STL-10 dataset, we use logistic regression to evaluate}
\vspace{0.5em}\centering
\begin{tabular}{ccccccccccccc}
\toprule[1.5pt]
method & epoch:0 &4&8&12&16&20&24&28&32\\
\midrule[1pt]
A-NDCG(train)&77.8&85.0&86.1&86.7&86.6&86.6&87.4&87.0&86.3\\
A-NDCG+mixup(train)&80.9&86.5&87.2&87.9&89.8&89.3&89.6&89.1&89.4\\
A-NDCG+cutmax(train)&88.4&86.4&87.1&87.5&87.6&87.3&87.8&87.7&88.3\\

A-NDCG(test)&44.0&49.5&51.9&52.3&51,8&52.1&52.8&50.9&52.4\\
A-NDCG+mixup(test)&44.9&52.0&52.7&53.1&53.8&54.3&53.8&54.8&55.7\\
A-NDCG+cutmax(test)&51.0&52.2&52.3&51.9&52.5&52.5&53.0&53.0&53.0\\

\bottomrule[1.5pt]
\end{tabular}

\label{table:self_mix_acc}
\end{table*}
\begin{table*}[htbp]
\caption{Classification accuracy on mixed STL-10 dataset, we use knn to evaluate}
\vspace{0.5em}\centering
\begin{tabular}{ccccccccccccc}
\toprule[1.5pt]
method & epoch:0 &4&8&12&16&20&24&28&32\\
\midrule[1pt]
A-NDCG(train)&62.2&66.1&66.6&67.5&68.3&67.4&67.8&67.9&68.9\\
A-NDCG+mixup(train)&64.8&68.1&68.9&69.7&68.8&70.5&70.3&71.3&70.8\\
A-NDCG+cutmax(train)&63.6&63.5&66.8&68.0&69.5&69.0&69.4&69.9&70.0  \\
A-NDCG(test)&36.3&40.5&42.4&43.0&43.0&43.2&44.0&44.8&44.7\\
A-NDCG+mixup(test)&39.2&43.6&44.3&46.2&47.0&46.9&47.5&46.5&47.5\\
A-NDCG+cutmax(test)&37.7&42.1&42.6&44.4&44.6&45.4&46.3&45.9&45.8\\
\bottomrule[1.5pt]
\end{tabular}
\vspace{\baselineskip}
\end{table*}
\subsubsection{Implementation Details}
For the experiments on the STL-10 dataset, we follow the approach of \cite{varamesh2020self}. We use ResNet-18 as the encoder, followed by a projection network composed of two layers of MLP. The batch size is set to 32, and the optimizer and learning rate are the same as those used in SimCLR. The number of epochs is set to 36, and longer training times yield better results.

\subsubsection{Experimental Results Analysis}
Tables \ref{table:self-acc} and \ref{table:self-knn} indicate that under the two evaluation algorithms, the performance of A-NDCG is better than SimCLR~\cite{chen2020simple}. This is because there is no limit to the number of perspectives of a single data point. From the perspective of contrastive learning, A-NDCG compares the difference between the sample feature similarity set that is not sorted according to the real ranking relationship and the sample feature similarity set that is sorted according to the real ranking relationship. It does not compare the difference between positive sample pair similarity and negative sample pair similarity, thus surpassing the conceptual constraints of positive and negative samples, and has broader meaning and applicability. 

Tables \ref{table:self_mix_acc} and \ref{table:self_mix_knn} show that A-NDCG can fully utilize label-level data augmentation methods to transform the training data, making full use of the information in the unsupervised data. Despite the fact that manually designed data augmentation methods may introduce a lot of noise and errors, the loose constraints of A-NDCG reduce the influence of noise. It can be seen that the improvement effect of A-NDCG is still very significant. Making full use of the various information in unsupervised training data is obviously necessary for unsupervised learning.

\section{Conclusion}

In this paper, the representation learning problem across multiple tasks is modeled as a ranking problem. By adopting the ranking problem as a unified perspective, we solve the representation learning problem under different tasks by optimizing a unified objective. We conducted extensive experiments across various learning tasks, including classification, retrieval, multi-label learning, regression, and self-supervised learning, to demonstrate the superiority of the approximate NDCG loss, thereby verifying the effectiveness and broad applicability of the representation learning by ranking framework.

Furthermore, under the self-supervised learning task, we applied data augmentation methods to transform the training data, which improved the performance of the approximate NDCG loss. This demonstrates that the framework can fully utilize the information in unsupervised data. Representation learning by ranking offers us a unified perspective and provides deeper insights into understanding and designing representation learning methods.

\bibliography{bib}
\bibliographystyle{icml2021}

\end{document}